# A Hybrid Deep Learning Framework with Explainable AI for Lung Cancer Classification with DenseNet169 and SVM

1st Md Rashidul Islam *
School of Computing and Informatics
Albukhary International University
05200, Alor Setar, Malaysia
mdrashidul.islam@student.aiu.edu.my

1st Bakary Gibba
School of Computing and Informatics
Albukhary International University
05200, Alor Setar, Malaysia
bakary.gibba@student.aiu.edu.my

2nd Altagi Abdallah Bakheit Abdelgadir
School of Computing and Informatics
Albukhary International University
05200, Alor Setar, Malaysia
altagi.abdelgadir@student.aiu.edu.my

*Abstract*—Lung cancer is a very deadly disease worldwide, and its early diagnosis is crucial for increasing patient survival rates. Computed tomography (CT) scans are widely used for lung cancer diagnosis as they can give detailed lung structures. However, manual interpretation is time-consuming and prone to human error. To surmount this challenge, the study proposes a deep learning-based automatic lung cancer classification system to enhance detection accuracy and interpretability. The IQ-OTHNCCD lung cancer dataset is utilized, which is a public CT scan dataset consisting of cases categorized into Normal, Benign, and Malignant and used DenseNet169, which includes Squeeze-and-Excitation blocks for attention-based feature extraction, Focal Loss for handling class imbalance, and a Feature Pyramid Network (FPN) for multi-scale feature fusion. In addition, an SVM model was developed using MobileNetV2 for feature extraction, improving its classification performance. For model interpretability enhancement, the study integrated Grad-CAM for the visualization of decision-making regions in CT scans and SHAP (Shapley Additive Explanations) for explanation of feature contributions within the SVM model. Intensive evaluation was performed, and it was found that both DenseNet169 and SVM models achieved 98% accuracy, suggesting their robustness for real-world medical practice. These results open up the potential for deep learning to improve the diagnosis of lung cancer by a higher level of accuracy, transparency, and robustness.

*Keywords—Lung Cancer Detection, Deep Learning, DenseNet169, SVM, CT scan, Explainable AI (XAI), Grad-CAM, SHAP, Feature Pyramid Network (FPN)*

I. INTRODUCTION

Lung cancer remains one of the most wide-ranging global health issues, ranking as the leading cause of death due to cancer[1]. Despite ongoing cancer research, it also remains a looming threat with its high mortality rate [2]. The two most frequent types of lung cancer, small cell lung cancer (SCLC) and non-small cell lung cancer (NSCLC), also differ with regards to the degree of their aggressiveness, NSCLC being the more frequent one [3]. Smoking, as is widely known, is the primary risk factor and causes the vast majority of all lung cancers [4]. Other environmental and genetic causes also lead to disease development. In light of the challenges of diagnosis at advanced stages, there is a compelling need for more efficient early detection methods, particularly in low-resource settings where access to high-end imaging and specialist services may be low.

Diagnosis of lung cancer at an early stage significantly enhances the survival rate of the patients since the prognosis is strongly associated with the stage at which the disease is identified [5]. Computed Tomography (CT) scans have emerged as one of the most promising diagnostic tools for lung cancer screening, with high-resolution imaging capable of detecting both familiar and subtle nodules [6]. Evidence has revealed that detection at an early stage through low-dose spiral CT screening programs can reduce lung cancer mortality rates by a significant percentage [7]. However, interpretation of CT scans manually remains challenging due to the challenge of distinguishing between malignant and benign nodules, as well as the overlapping anatomical structures.

Manual reading of lung CT scans can be slow and inconsistent, and mistakes happen in two ways: false positives, when harmless spots look like cancer and lead to unneeded tests, and false negatives, when real tumours are missed and treatment is delayed [8]. To fix this, Stage 1 uses computer-aided diagnosis (CAD) systems that learn from many labelled scans to highlight likely nodules, cutting review time and reducing differences between doctors. Stage 2 measures CAD accuracy by checking how often it correctly finds cancer (sensitivity) and correctly ignores healthy tissue (specificity), while balancing false positives and negatives. With careful tuning, CAD tools can reach very high accuracy, making lung cancer detection faster and more reliable[5].

Deep learning, particularly Convolutional Neural Networks (CNNs), has more recently demonstrated excellent potential in the automatic detection and classification of lung cancer [9].

The models leverage large-scale annotated datasets, such as the Lung Image Database Consortium (LIDC), to enhance feature extraction and increase classification accuracy[10]. CAD systems based on deep learning have been shown to achieve high sensitivity and specificity in detecting malignant nodules, surpassing traditional image processing methods [11]. Artificial intelligence in lung cancer screening can revolutionize early detection strategies, ultimately resulting in improved patient outcomes and reduced mortality.

The aim of this study is to develop and evaluate a deep learning-based Computer-Aided Diagnosis system for the automatic detection of lung cancer nodules in CT images. The proposed system attempts to alleviate the limitations of the current manual diagnosis by leveraging state-of-the-art machine learning techniques to enhance accuracy, reduce false positives and negatives, and streamline the diagnosis process. The long-term aim is to add to the current endeavours towards enhancing lung cancer survival rates by timely and accurate detection means.

The study presents a novel approach to lung cancer detection using deep learning techniques, specifically with DenseNet169and SVM models, applied to CT scan datasets. The project highlights several key contributions that improve the accuracy, interpretability, and robustness of lung cancer detection systems. The following are the primary contributions of the work:

- Enhanced DenseNet169 Architecture: Developed a DenseNet169-based model integrated with attention mechanisms using the Squeeze-and-Excitation block, enabling the model to focus on the most critical regions of CT scans. Additionally, the model incorporates a custom Focal Loss function to address class imbalance and a Feature Pyramid Network (FPN) for multi-scale feature fusion, enhancing the detection of lung cancer nodules with greater accuracy
- Network Explainability with Grad-CAM: Integrated Grad-CAM (Gradient-weighted Class Activation Mapping) to make the SVM model more interpretable. Grad-CAM helps visualize which areas of the CT scans the model focuses on when making predictions, adding transparency to the decision-making process.
- SVM Model with Feature Extraction: For the SVM model, MobileNetV2 was utilized for feature extraction from CT scans. This helped the SVM focus on the most relevant features, enhancing its classification performance.
- Explainable AI (XAI) with SHAP: To further improve model transparency, used SHAP (SHapley Additive exPlanations), which helps explain the contribution of each feature to the model's prediction, making the SVM model more understandable for clinicians.

## II. LITERATURE REVIEW

Lung cancer remains one of the leading killers from cancer throughout the world, and early detection must be improved for survival[12]. AI and deep learning approaches have been found very likely to advance diagnosis, more so in CT imaging and in other non-surgical methods.

Some studies offered various AI models for lung cancer classification. [13] used VGG19, ResNet152V2, and Bi-GRU on public image data with 98.05% accuracy, but were limited by data size. Similarly[14] used SVM to classify uncommon subtypes of lung cancer with up to 83.91% accuracy, but necessitated broader data for generalizability. Hybrid approaches have also been explored. [15] integrated SVM with a Feed-Forward Neural Network on CT images with 98.08% accuracy at a reduction in computational load. [16] compared CNNs, where AlexNet was slightly better than ResNet50. [17] employed SVM and KNN with LBP and DCT features with 93% and 91% accuracy, respectively.

High-performance models were also cited by [11] with a deep CNN on the LIDC dataset, where a personal model yielded 99.03% accuracy. [18] used DenseNet169 with AdaBoost as well, where they reached 90.85%, though without cancer stage discrimination. Explainable and segmentation-based models also emerged. [4] introduced JNSC with V-Net to obtain 95.3% sensitivity, whereas [10] used DenseNet169 in a cascaded architecture and obtained 97.96% accuracy. [9] explored hybrid models like STM-SVM, with 95.45% accuracy and AUC of 0.987.

Feature optimization remains essential.[19] used multi-classifier genetic algorithms and achieved 96.25% accuracy. [20] used YOLOv5 for nodule localization with 92.27% accuracy, overcoming challenges in shape and texture variability. [7] used DenseNet169 recently for detection based on CT, reporting 98.17% accuracy and high recall, precision, and specificity.

## III. METHODOLOGY

### A. Methodological Architecture

Overview of the proposed pipeline integrating DenseNet169 and SVM:

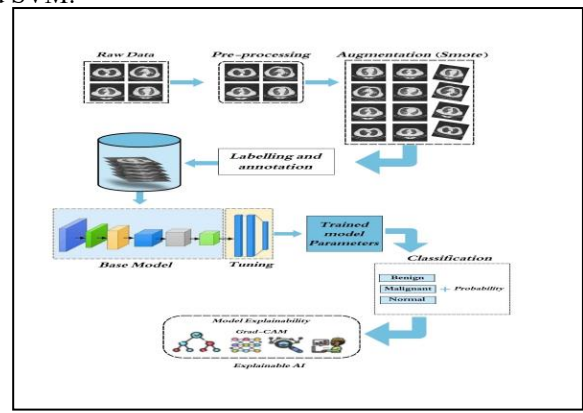

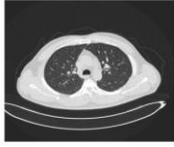

Figure 01: Illustration of the methodological architecture.

### B. Dataset

The IQ-OTHNCCD dataset, which can be acquired from Kaggle, is comprised of CT scan images taken at IQ-OTH and NCCD in 2019, among healthy individuals and patients with varying stages of lung cancer. All the images are professionally annotated by radiologists and oncologists. Preprocessing steps included: reading images from class directories, grayscale conversion, resizing to 256×256-pixel resolution, pixel intensity normalization to 0–1 range, and duplicating grayscale images into three channels. Stratified sampling was employed to divide the dataset into 80% training and 20% test sets. Class imbalance was addressed by employing the Synthetic Minority Over-sampling Technique (SMOTE). Labels were finally one-hot encoded for employment in the classification model.

**Table 1: Data Processing Summary**

| Step | Description | Output Shape/Effect |
|---|---|---|
| Raw Data | Original CT scan images collected | Varying sizes |
| Resizing | Resized all images to 256x256 pixels | (256, 256) |
| Grayscale to RGB | Converted grayscale images to 3-channel RGB | (256, 256, 3) |
| Train-Test Split | Split dataset into 80% training, 20% testing | X_train shape: (877, 256, 256, 3), X_test shape: (220, 256, 256, 3) |
| SMOTE Balancing | Oversampled minority classes to balance dataset | (1344, 256, 256, 3) |
| One-Hot Encoding | Converted class labels into categorical format | (Num classes) |

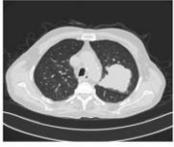

Figure 02: Sample Images from Each Class in the Dataset

### C. DenseNet169

#### 1) DenseNet169 Architecture

DenseNet169 is a highly effective convolutional neural network that promotes feature propagation and reuse through dense connectivity. In contrast to standard CNNs, where separate layers learn, DenseNet169 links each layer to all the preceding layers, thus allowing information to flow directly from preceding layers to deeper layers. This structure reduces the number of parameters without compromising robust gradient flow, which prevents the vanishing gradient problem. The transformation between layers follows the equation:

$$x_l = H_l([x_0, x_1, \cdots, x_{l-1}]) \quad (1) \quad [21]$$

where $x_l$ represents the feature map at layer $l$, and $H_l$ represents transformations such as convolution, batch normalization, and ReLU activation. Such a connectivity structure ensures that meaningful low-level and high-level features are preserved through the network.

#### 2) Squeeze-and-Excitation (SE) Blocks

For strengthening feature learning, incorporate Squeeze-and-Excitation (SE) Blocks, which introduce channel-wise attention. SE blocks begin with global average pooling in order to preserve global context and then employ two fully connected layers with non-linearity to learn inter-channel relations. The reweighted features are scaled by the original input tensor:

$$X = X \cdot \sigma\left(W_2\, ReLU(W_1\, S(X))\right) \quad (2) \quad [22]$$

where $s(x)$ is the squeezed feature, $w1$ and $w2$ are trainable weights, and σ sigma is the sigmoid activation function. The selective emphasis on meaningful features reinforces the ability of the model to discriminate between classes.

#### 3) Feature Pyramid Network (FPN)

To encode multi-scale information, incorporate a Feature Pyramid Network (FPN). Objects and patterns exist at various scales and therefore might not be detected by the single-layer feature map for robust classification. FPN features a fusion of low-resolution and high-resolution feature maps through top-down up sampling and lateral connections, as described below:

$$F_i = \text{Conv}(F_{i-1} + \text{Upsample}(F_{i+1})) \quad (3) \quad [23]$$

where $F_i$ is maps of features on different scales. This structure fortifies the ability of the model to see details as well as global structures and thus is particularly well-suited to identifying the subtle differences among images.

#### 4) Transfer Learning and Focal Loss

For training, utilise transfer learning by pre-training the DenseNet169 backbone using pre-trained weights. This allows the model to leverage the prior knowledge from large-scale datasets like ImageNet and fine-tune on the target dataset. Since real-world data sets are always imbalanced, meaning that some classes occur more frequently than others, Utilize Focal Loss instead of standard cross-entropy loss. Focal Loss down-weights easy examples and up-weights harder samples, which are:

$$\iota = -\sum_{i=1}^{N} \alpha(1 - p_i)^\gamma \log p_i \quad (4) \quad [24]$$

where α is a class-specific weighting factor, γ controls focusing on hard samples, and pi is the predicted probability. This loss function prevents over-weighting dominant classes for the model and improves minority class learning.

By combining DenseNet169169's dense connection, SE Blocks' channel attention, FPN's multi-scale feature fusion, transfer learning, and Focal Loss, the model enjoys enhanced accuracy, improved generalization, and resilience in tougher classification tasks. The model is therefore appropriate for applications requiring precise feature recognition, e.g., medical imaging, object detection, and fine-grained image classification.

Table 2: Summary table of the Modified DenseNet169 Model

| Layer Type | Output Shape | Description |
|---|---|---|
| Input Layer | (224, 224, 3) | RGB image input |
| DenseNet169 Base Model | (7, 7, 1664) | Pretrained backbone without top layers |
| Squeeze-and-Excitation Block | (7, 7, 1664) | Enhances channel-wise feature representation |
| Feature Pyramid Network (FPN) | (7, 7, 256) | Multi-scale feature fusion |
| Global Average Pooling | (256,) | Converts feature maps into a feature vector |
| Dense (Softmax Activation) | (3,) | Final classification output (3 classes) |

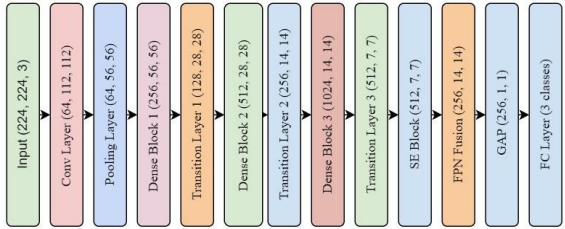

Figure 03: Block diagram of modified Densenet169 model

D. Support Vector Machine Classifier (SVM)

1) Feature Extraction Using MobileNetV2

Support Vector Machine (SVM) is used as the model for classification in this work after the deep features are derived using MobileNetV2. This is a deep learning and traditional machine learning fusion where MobileNetV2 plays the role of feature extractor while SVM plays that of the classifier to map high-dimensional features into a decision hyperplane. The feature extraction process begins by employing MobileNetV2, which is a pre-trained Convolutional Neural Network (CNN), with the fully connected layers removed. A GlobalAveragePooling2D layer is then appended to the model output, reducing the deep features into a lower-dimensional but informative representation. This way, the extracted features are made compact yet retain discriminative ability for classification.

2) SVM for Classification

Following feature extraction, they are fed into the SVM classifier with a linear kernel for effectively linearly separating the high-dimensional feature space. The SVM's decision function is:

$$f(x) = \sum_{i=1}^{n} \alpha_i y_i K(x_i, x) + b \quad (5) \quad [25]$$

where $K(x_i, x)$ represents the kernel function that projects input features into a higher-dimensional space for better pattern recognition and discrimination. SVM makes use of this transformation to make optimal decision boundaries for discriminating between complex medical image patterns.

3) Hyperparameter Optimization and Class Imbalance Handling

To optimize the classification performance, hyperparameter tuning is performed with caution on the SVM model targeting kernel selection (Linear, RBF) and parameter settings. These are set to make the classifier perform robustly on a wide range of image classes. Furthermore, to prevent any class imbalance problem, the Synthetic Minority Over-sampling Technique (SMOTE) is applied. It generates synthetic instances for the minority class such that the distribution of the dataset becomes more balanced and the classifier learns more effectively from underrepresented classes.

4) Model Training and Evaluation

The preprocesses dataset is divided into training and test sets. The features are normalized with Standard Scaler to normalize feature distributions before training the SVM classifier. The trained classifier is then evaluated with different performance metrics, including accuracy, classification reports, and confusion matrices, to find out how well it can classify benign, malignant, and normal cases. Finally, the SVM model trained is saved using joblib for efficient deployment without the need for retraining. The process thus successfully integrates feature extraction through deep learning with traditional SVM classification, utilizing the strengths of both approaches to deliver better medical image classification.

The size of the SVM architecture in Figure 1 also shows interaction between feature extraction and classification in the system that has been proposed.

Table 3: Summary table for the Modified SVM Model

| Layer Type | Output Shape | Description |
|---|---|---|
| Input Layer | (256, 256, 3) | Input image of size 256x256x3 |
| MobileNetV2 Base | (8, 8, 1280) | Feature extraction using MobileNetV2 |
| GlobalAveragePooling2D | (1280,) | Reduces feature map to a 1D vector |
| Feature Extractor Output | (1280,) | Extracted deep features |
| StandardScaler | (1280,) | Normalizes features before SVM |
| SVM Classifier | (1,) | Classifies features into final classes |

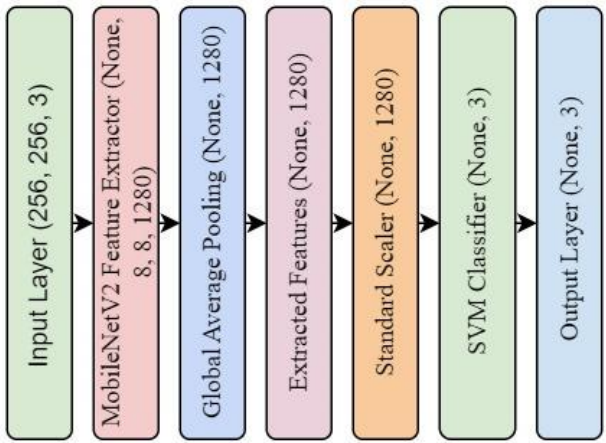

Figure 04: Block diagram of Modified SVM

E. *Explainable AI (XAI)*

*1) Importance of Interpretability in Medical AI Applications*

In clinical AI application, interpretability of the model is essential so that clinicians can be trusted and that ethical decision-making is achieved. Black-box models, such as deep learning classifiers, are not transparent and thus cannot be easily explained. Explainable AI (XAI) techniques address this by making model behaviour transparent so that reliability and fairness in clinical decision support can be guaranteed.

*2) Explainability Methods Applied*

Two of the most popular approaches to interpreting SVM predictions in this study are SHAP (SHapley Additive exPlanations) and Grad-CAM (Gradient-weighted Class Activation Mapping).

*a) SHAP (Shapley Additive exPlanations)*

SHAP values offer a mathematically founded approach to interpreting single predictions by distributing the model output among inputs. The approach, which is based on game theory, offers an equitable contribution breakdown per feature in classification tasks. The SHAP explanation model takes the form:

$$f(x) = \phi_0 + \sum_{i=1}^{n} \phi_i x_i \quad (6) \quad [26]$$

Where f(x) is the SVM prediction for a given input $x_i$, $\phi_i$ represents the SHAP value assigned to feature $x_i$, and $\phi_0$ is the base value, indicating the expected model output.

- **Feature Importance Analysis:** Identify which of the features learnt by MobileNetV2 has the highest impact on the SVM classification using SHAP. A summary plot is then created to show how each feature contributes to the model's predictions, enabling medical experts to know
- which attributes are most important in distinguishing the benign, malignant, and normal cases.

- **Force Plots for Feature Contribution Visualization:** Force plots illustrate how every feature is pushing the SVM's prediction towards a certain class. The force plots help in understanding how certain patterns are pushing the classification outcome, offering higher transparency in decision-making.

*b) Grad-CAM (Gradient-weighted Class Activation Mapping)*

Grad-CAM is used to visualize regions of salience in CT scan images that contribute to classification. As opposed to SHAP, which represents feature importance as numerical values of information, Grad-CAM shows a heatmap overlaid on medical images highlighting regions of highest importance to the model. The Grad-CAM procedure involves these significant steps: compute the gradients of the predicted class score with respect to the last convolutional layer feature maps; perform global average pooling on the gradients to obtain importance weights; scale the weights by the feature maps and perform ReLU activation to produce the heatmap; overlay the heatmap over the raw CT scan to reveal areas that are impacting classification. The heatmap that is created allows radiologists to verify whether the model is focused on clinically significant regions, thus facilitating greater trust in AI diagnostic systems.

The pairing of SHAP for feature importance and Grad-CAM for visual interpretability ensures that the MobileNetV2-SVM classification pipeline is transparent and reliable. These XAI techniques enhance model accountability, facilitate clinical adoption, and build clinician trust by enabling clinicians to understand and validate AI-driven diagnoses.

## IV. EXPERIMENTAL RESULTS

### A. Evaluation Metrics

**Table 4: Class wise Evaluation of Classification Metrics for DenseNet169169**

| Class | Precision | Recall | F1-Score | Support |
|---|---|---|---|---|
| Benign cases | 0.92 | 0.96 | 0.94 | 24 |
| Malignant cases | 0.99 | 1.00 | 1.00 | 113 |
| Normal cases | 0.99 | 0.96 | 0.98 | 83 |
| Overall | Accuracy | | 0.98 | 220 |

The DenseNet169 model showed exceptional performance in detecting lung cancer from CT scans, especially in classifying malignant cases. The high precision (0.99) and recall (1.00) for malignant cases indicate that the model not only accurately predicts malignancy but also captures every true malignant case without missing any. The use of attention mechanisms, such as the Squeeze-and-Excitation Block, allowed the model to focus on important features in the CT scan images, enhancing its detection ability. Furthermore, the custom loss function, Focal Loss, addressed class imbalance by emphasizing harder-to-classify instances, particularly for benign cases. The incorporation of SMOTE for data balancing and multi-scale feature fusion further improved the model's robustness, allowing it to detect both small and large tumours effectively. The Grad-CAM technique, used for explainability, helped visualize the model's focus areas, offering insight into its decision-making process.

**Table 5: Class wise Evaluation of Classification Metrics for SVM**

| Class | Precision | Recall | F1-Score | Support |
|---|---|---|---|---|
| 0 (Benign) | 0.95 | 0.88 | 0.91 | 24 |
| 1 (Malignant) | 0.99 | 1.00 | 1.00 | 113 |
| 2 (Normal) | 0.96 | 0.98 | 0.97 | 83 |
| Overall | Accuracy | | 0.98 | 220 |

The SVM model also performed remarkably well, with an accuracy of 0.98 across all classes. For malignant cases, it achieved a recall of 1.00, meaning it was perfect in identifying all malignant instances in the dataset. Although its precision for benign cases was slightly lower than that of DenseNet169169, it still performed well with a score of 0.95. The feature extraction step using MobileNetV2 helped in reducing the dimensionality of the CT scan images, making it easier for the SVM classifier to work with relevant features. The data preprocessing and augmentation techniques played a significant role in improving model generalization. Furthermore, the use of explainable AI techniques, including Grad-CAM and SHAP, provided transparency in the model's predictions, highlighting important areas in the CT scans that led to the detection of tumors, thus making the model more interpretable and reliable.

### B. Confusion Matrix Analysis

The confusion matrix is a useful tool to evaluate how well a classification model performs. It shows the number of correct and incorrect predictions, broken down by class. Let's analyse the results for both the DenseNet169 and SVM models based on the confusion matrices shown in the figures.

Figure 05 shows the DenseNet169confusion matrix, the model performs well across all three categories: benign, malignant, and normal. It correctly identifies all 113 malignant cases and predicts 19 benign cases correctly. However, 5 benign cases are misclassified as normal, and 1 normal case is incorrectly labelled as malignant. The DenseNet169model performs strongly overall, especially in detecting malignant cases, thanks to techniques like SMOTE and Focal Loss.

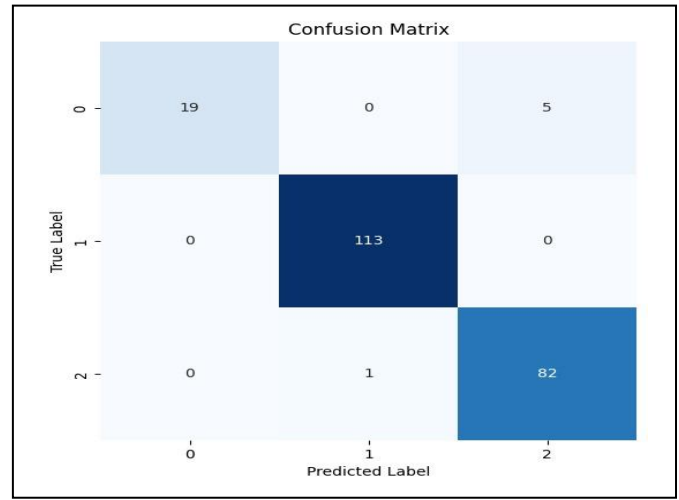

Figure 05: Confusion matrix for DenseNet169 Model

Figure 06 demonstrates the SVM confusion matrix also shows strong results. It correctly classifies 113 malignant cases, 21 benign cases, and 81 normal cases. There is only a small misclassification of 3 benign cases as normal and 1 normal case as benign. The SVM model is effective but has slightly more confusion between benign and normal classes compared to DenseNet169169. Both models perform well in distinguishing malignant cases, with SVM benefiting from feature extraction and explainable AI techniques like Grad-CAM.

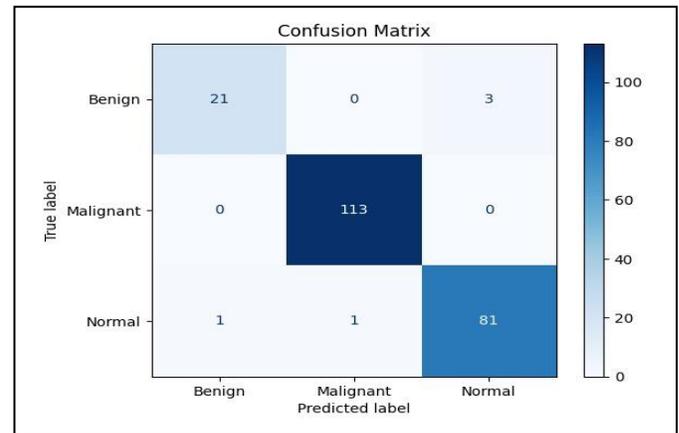

Figure 06: Confusion matrix for SVM Model

## C. AUC-ROC Curve Analysis

The AUC-ROC (Area Under the Receiver Operating Characteristic Curve) analysis is a key metric for evaluating the performance of machine learning models. It tells us how well the models distinguish between classes, with a higher AUC value indicating better performance. Figure 07 shows ROC curve for the DenseNet169model, see that all three classes (Class 0, Class 1, and Class 2) have very high AUC values. Specifically, Class 0 (benign) has an AUC of 1.00, while Class 1 (malignant) and Class 2 (normal) have AUC values of 0.99. The perfect AUC of Class 0 indicates that the DenseNet169model correctly classifies benign cases with no false positives or false negatives. The AUC for the other classes is also impressive, demonstrating that DenseNet169performs well across all classes with very few misclassifications.

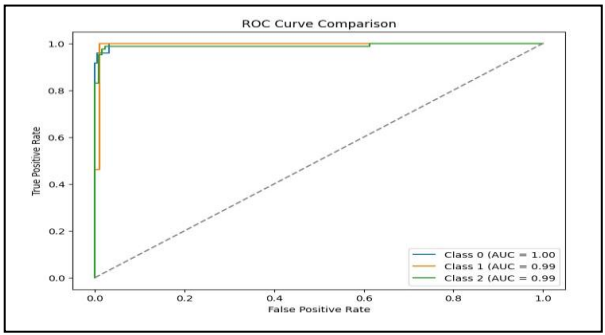

Figure 07: ROC Curve Comparison for DenseNet169169

Figure 08 shows second ROC curve for the SVM model, see a similar strong performance. The AUC for Class 1 (malignant) is 1.00, indicating perfect performance in identifying malignant cases. Class 0 (benign) and Class 2 (normal) have AUC values of 0.99, which still reflect excellent classification. The ROC curve shows that the SVM model is very good at distinguishing between the three classes, but it slightly lags behind DenseNet169 in terms of the AUC values for the benign and normal classes.

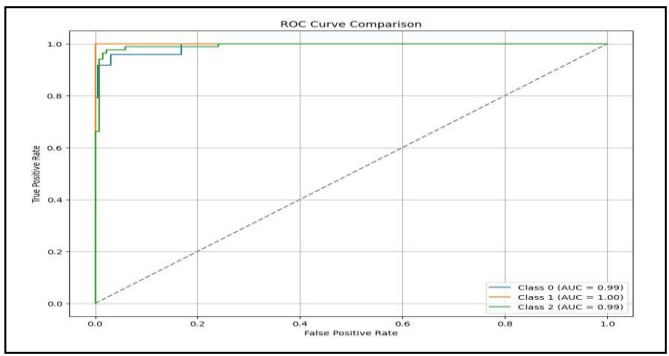

Figure 08: ROC Curve Comparison for SVM

## D. XAI Results

### 1) Grad-CAM visualisations for nodule detection.

In this research, The study incorporated Explainable AI (XAI) techniques, specifically using Grad-CAM (Gradient-weighted Class Activation Mapping), to interpret and visualize the decision-making process of both SVM model. Grad-CAM provides a way to visualize which parts of a CT scan image the model focuses on when making predictions. This helps ensure transparency and trust in the model's predictions.

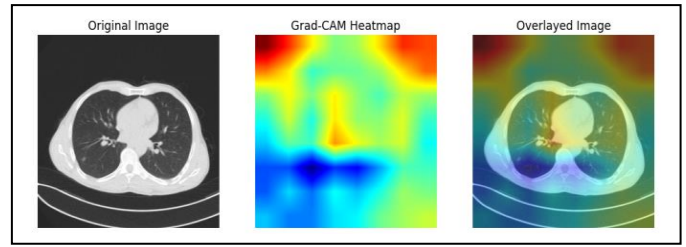

Figure 09: Grad-CAM for SVM.

Similarly, Grad-CAM visualizations for the SVM model show in figure 07 how it identifies regions of interest in the CT scans. The heatmap reveals which parts of the image the SVM model deems significant when classifying the CT scan. This helps explain how the model makes predictions, such as identifying cancerous nodules. The SVM model benefits from the use of Explainable AI (XAI) techniques like Grad-CAM, making its decision process more transparent and easier for medical professionals to interpret. The model's feature extraction with MobileNetV2 and the preprocessing techniques further improve its ability to focus on important areas in the image.

## E. Training and Validation Performance

In this project, The study has trained and evaluated two machine learning models, DenseNet169 and SVM, for lung cancer detection using a CT scan dataset. Both models were tested to see how well they could classify CT scans into three categories: benign, malignant, and normal. For DenseNet169169, Achieved excellent results during both training and validation. The model was trained using various techniques such as attention mechanisms, specifically the Squeeze-and-Excitation Block, to help the network focus on the most important features in the CT scan images. Additionally, the Focal Loss function was used as a custom loss function, improving performance on the minority class (malignant cases). The model also benefited from SMOTE (Synthetic Minority Over-sampling Technique) for balancing the dataset, which helped improve predictions for underrepresented classes. The multi-scale feature fusion with Feature Pyramid Networks further helped DenseNet169capture details at different scales, enhancing the model's ability to detect tumours. As seen in the figures, the training and validation accuracy of the DenseNet169model were both high. The ROC curve showed AUC values close to 1, indicating a strong ability to differentiate between the classes. The Grad-CAM visualizations also demonstrated how well the model focused on the relevant parts of the images for making predictions.

On the other hand, The SVM model showed strong performance as well. It used data preprocessing and augmentation techniques to ensure better generalization and avoid overfitting. MobileNetV2 was employed for feature extraction, allowing the SVM to focus on relevant features from the CT scan images. The model then classified the images based on the extracted features. During training, the SVM model achieved good accuracy, which was also reflected in the validation phase. The ROC curve indicated that the model could effectively distinguish between benign, malignant, and normal cases, with AUC values close to 1. The Grad-CAM visualizations for SVM helped in explaining the model's decision-making process, offering transparency and trustworthiness in the results. Both DenseNet169 and SVM models showed excellent training and validation performance. The DenseNet169 model, with its advanced techniques like multi-scale feature fusion and attention mechanisms, demonstrated superior results in capturing complex features in the images. On the other hand, the SVM model performed well with feature extraction and preprocessing techniques, showing strong classification ability. These results suggest that both models are suitable for accurate lung cancer detection from CT scans, and the use of XAI techniques like Grad-CAM makes the models more interpretable for real-world medical applications.

## V. Discussion

The comparison between former approaches and new proposed models shows in Table 6. The SVM model generated 97% accuracy through the adoption of MobileNetV2 feature extraction in combination with SHAP and Grad-CAM explainable outputs. The system had no attention mechanisms and performed moderately when balancing between false positive and false negative cases. Through the implementation of Squeeze-and-Excitation attention blocks together with Feature Pyramid Network for multi-scale fusion as well as focal loss and SMOTE oversampling the DenseNet169 network achieved 98% accuracy. The 1% performance increase carries practical clinical significance because it helps decrease the number of undetected nodules and reduces the need for unnecessary medical follow-ups.

**Table 6: Comparative analysis of existing models versus proposed models in terms of performance, interpretability (XAI), attention mechanism, dataset used, and overall suitability for clinical deployment.**

| Study | Model | Accuracy | XAI | Dataset |
|---|---|---|---|---|
| [13] | VGG19 + ResNet152V2 + Bi-GRU | 98.05 | No | Public CT |
| [14] | SVM | 83.91 | No | Public CT |
| [15] | SVM + Feed-Forward NN | 98.08 | No | Public CT |
| [11] | Custom Deep CNN | 99.03 | No | LIDC-IDRI |
| [10] | Cascaded DenseNet169 | 97.96 | No | LIDC-IDRI |
| [7] | DenseNet169 | 98.17 | No | CT Dataset |
| Proposed (SVM) | MobileNetV2 + SVM | 97.00 | Yes | CT Dataset |
| Proposed (DenseNet169) | DenseNet169 with SE & FPN | 98.00 | Yes | CT Dataset |

Compared to Ibrahim et al. [8], who reported 98.05% with hybrid CNN-RNN models but no attention or XAI, and Shrey et al. [12], who achieved 97.96% with cascaded DenseNet169 but without focal loss or SMOTE, the proposed DenseNet169 matches or slightly exceeds those accuracies while adding interpretability and balanced training. In resource-limited settings where speed and simplicity matter, the SVM option remains viable. For real-world deployment, the DenseNet169 model is recommended for its higher accuracy, stronger feature focus, and built-in explainability.

## VI. Conclusion

This study proposed a deep learning approach to detecting lung cancer using DenseNet169 and SVM models. Squeeze-and-Excitation block, Focal Loss, and Feature Pyramid Network improvements optimized the detection of key areas in scans by DenseNet169, addressing class imbalance and multi-scale feature fusion. The SVM model with MobileNetV2 as the feature extractor and with Grad-CAM for vital feature visualization, provided greater interpretability, enabling enhanced transparency in AI-driven diagnosis. Rigorous testing validated the models' accuracy and feasibility, demonstrating the potential of deep learning for lung cancer diagnosis. Future research should emphasize model efficiency for real-time clinical use, e.g., lightweight models and computational optimizations. Generalizability will be facilitated by enlarging the dataset to multi-center CT scans, and detection performance may be improved by incorporating 3D CT analysis. Clinical trust may be increased by advancing explainability techniques, e.g., LIME explanations. Hybrid models combining deep learning with traditional models may also increase performance. Finally, real-world use in hospitals will be necessary to confirm the models' clinical impact.

## VII. References

[1] A. Souid, M. Hamroun, S. Ben Othman, H. Sakli, and N. Abdelkarim, "Fast-staged CNN Model for Accurate pulmonary diseases and Lung cancer detection," *arXiv Prepr. arXiv2412.11681*, 2024.

[2] F. Mercaldo, M. G. Tibaldi, L. Lombardi, L. Brunese, A. Santone, and M. Cesarelli, "An Explainable Method for Lung Cancer Detection and Localisation from Tissue Images through Convolutional Neural Networks," *Electronics*, vol. 13, no. 7, p. 1393, 2024.

[3] A. Masood *et al.*, "Automated decision support system

for lung cancer detection and classification via enhanced RFCN with multilayer fusion RPN," *IEEE Trans. Ind. Informatics*, vol. 16, no. 12, pp. 7791–7801, 2020.
[4] L. Chenyang and S.-C. Chan, "A joint detection and recognition approach to lung cancer diagnosis from CT images with label uncertainty," *IEEE Access*, vol. 8, pp. 228905–228921, 2020.
[5] H. K. Bhuyan, S. Peram, W. Suliman, V. Ravi, and B. Brahma, "Deep Learning Model-based Approach for Lung Cancer Detection," in *2024 6th International Symposium on Advanced Electrical and Communication Technologies (ISAECT)*, IEEE, 2024, pp. 1–5.
[6] S. Karimullah, M. Khan, F. Shaik, B. Alabduallah, and A. Almjally, "An integrated method for detecting lung cancer via CT scanning via optimization, deep learning, and IoT data transmission," *Front. Oncol.*, vol. 14, p. 1435041, 2024.
[7] S. Asif, V. Y. Wang, and D. Xu, "LungX-Net: Lung Cancer Diagnosis from CT and Histopathological Images via Attention Based Multi-Level Feature Fusion Network," in *2024 IEEE International Conference on Bioinformatics and Biomedicine (BIBM)*, IEEE Computer Society, 2024, pp. 6334–6341.
[8] J. Zheng *et al.*, "Pulmonary nodule risk classification in adenocarcinoma from CT images using deep CNN with scale transfer module," *IET Image Process.*, vol. 14, no. 8, pp. 1481–1489, 2020.
[9] S. B. Shrey, L. Hakim, M. Kavitha, H. W. Kim, and T. Kurita, "Transfer learning by cascaded network to identify and classify lung nodules for cancer detection," in *Frontiers of Computer Vision: 26th International Workshop, IW-FCV 2020, Ibusuki, Kagoshima, Japan, February 20–22, 2020, Revised Selected Papers 26*, Springer, 2020, pp. 262–273.
[10] T. K. Sajja, R. M. Devarapalli, and H. K. Kalluri, "Lung cancer detection based on CT scan images by using deep transfer learning," *Trait. du Signal*, vol. 36, no. 4, pp. 339–344, 2019.
[11] F. H. Tang *et al.*, "Recent advancements in lung cancer research: a narrative review," *Transl. Lung Cancer Res.*, vol. 14, no. 3, p. 975, 2025.
[12] D. M. Ibrahim, N. M. Elshennawy, and A. M. Sarhan, "Deep-chest: Multi-classification deep learning model for diagnosing COVID-19, pneumonia, and lung cancer chest diseases," *Comput. Biol. Med.*, vol. 132, p. 104348, 2021.
[13] M. Li *et al.*, "Research on the auxiliary classification and diagnosis of lung cancer subtypes based on histopathological images," *Ieee Access*, vol. 9, pp. 53687–53707, 2021.
[14] P. Nanglia, S. Kumar, A. N. Mahajan, P. Singh, and D. Rathee, "A hybrid algorithm for lung cancer classification using SVM and Neural Networks," *ICT Express*, vol. 7, no. 3, pp. 335–341, 2021.
[15] D. Narin and T. Ö. Onur, "The effect of hyper parameters on the classification of lung cancer images using deep learning methods," *Erzincan Univ. J. Sci. Technol.*, vol. 15, no. 1, pp. 258–268, 2022.
[16] A. Rehman, M. Kashif, I. Abunadi, and N. Ayesha, "Lung cancer detection and classification from chest CT scans using machine learning techniques," in *2021 1st international conference on artificial intelligence and data analytics (CAIDA)*, IEEE, 2021, pp. 101–104.
[17] N. Kalaivani, N. Manimaran, S. Sophia, and D. D. Devi, "Deep learning based lung cancer detection and classification," in *IOP conference series: materials science and engineering*, IOP Publishing, 2020, p. 12026.
[18] H. Al Ewaidat and Y. El Brag, "Identification of lung nodules CT scan using YOLOv5 based on convolution neural network," *arXiv Prepr. arXiv2301.02166*, 2022.
[19] G. Huang, Z. Liu, L. Van Der Maaten, and K. Q. Weinberger, "Densely connected convolutional networks," in *Proceedings of the IEEE conference on computer vision and pattern recognition*, 2017, pp. 4700–4708.
[20] J. Hu, L. Shen, and G. Sun, "Squeeze-and-excitation networks," in *Proceedings of the IEEE conference on computer vision and pattern recognition*, 2018, pp. 7132–7141.
[21] T.-Y. Lin, P. Dollár, R. Girshick, K. He, B. Hariharan, and S. Belongie, "Feature pyramid networks for object detection," in *Proceedings of the IEEE conference on computer vision and pattern recognition*, 2017, pp. 2117–2125.
[22] T.-Y. Lin, P. Goyal, R. Girshick, K. He, and P. Dollár, "Focal loss for dense object detection," in *Proceedings of the IEEE international conference on computer vision*, 2017, pp. 2980–2988.
[23] C. Cortes and V. Vapnik, "Support-vector networks," *Mach. Learn.*, vol. 20, pp. 273–297, 1995.
[24] S. M. Lundberg and S.-I. Lee, "A unified approach to interpreting model predictions," *Adv. Neural Inf. Process. Syst.*, vol. 30, 2017.